\documentclass[english,pre,onecolumn,superscriptaddress,notitlepage,nofootinbib,citautoscript]{revtex4-1}
\expandafter\let\csname equation*\endcsname\relax
\expandafter\let\csname endequation*\endcsname\relax
\usepackage[tbtags]{amsmath}
\usepackage{amsthm,amssymb,dsfont,mathrsfs,bbm,xfrac,tabularx,booktabs}
\usepackage{mathtools}
\usepackage{multirow}
\usepackage{upgreek}
\usepackage[usenames,dvipsnames]{xcolor}
\usepackage[caption=false]{subfig}
\usepackage[english]{babel}
\usepackage{pgfplotstable}
\usepackage{hyperref}
\hypersetup{pdfauthor={Baldassi Malatesta Negri Zecchina},pdftitle={},%
            colorlinks, linktocpage=true, pdfstartpage=3, pdfstartview=FitV,%
    breaklinks=true, pdfpagemode=UseNone, pageanchor=true, pdfpagemode=UseOutlines,%
    plainpages=false, bookmarksnumbered, bookmarksopen=true, bookmarksopenlevel=1,%
    hypertexnames=true, pdfhighlight=/O,%
    urlcolor=orange, linkcolor=blue, citecolor=red, 
        }

\usepackage{lmodern}
\usepackage[T1]{fontenc}
\usepackage[cal=boondoxo]{mathalfa}

\usepackage{graphicx}
\usepackage{grffile}

\catcode`,\active

\catcode`\,12

\makeatletter
\newsavebox{\@brx}
\newcommand{\llangle}[1][]{\savebox{\@brx}{\(\m@th{#1\langle}\)}%
  \mathopen{\copy\@brx\kern-0.5\wd\@brx\usebox{\@brx}}}
\newcommand{\rrangle}[1][]{\savebox{\@brx}{\(\m@th{#1\rangle}\)}%
  \mathclose{\copy\@brx\kern-0.5\wd\@brx\usebox{\@brx}}}
\makeatother

\allowdisplaybreaks
\begin{document}
\title[]{Wide flat minima and optimal generalization in classifying high-dimensional Gaussian mixtures}

\author{Carlo Baldassi}
\affiliation{Artificial Intelligence Lab, Bocconi University, 20136 Milano, Italy}
\author{Enrico M. Malatesta}
\affiliation{Artificial Intelligence Lab, Bocconi University, 20136 Milano, Italy}
\author{Matteo Negri}
\affiliation{Artificial Intelligence Lab, Bocconi University, 20136 Milano, Italy}
\affiliation{Dept. Applied Science and Technology, Politecnico di Torino, Corso Duca degli Abruzzi 24, I-10129 Torino, Italy}
\author{Riccardo Zecchina}
\affiliation{Artificial Intelligence Lab, Bocconi University, 20136 Milano, Italy}

\begin{abstract}
We analyze the connection between minimizers with good generalizing properties and high local entropy regions of a threshold-linear classifier in Gaussian mixtures with the mean squared error loss function. We show that there exist configurations that achieve the Bayes-optimal generalization error, even in the case of unbalanced clusters. We explore analytically the error-counting loss landscape in the vicinity of a Bayes-optimal solution, and show that the closer we get to such configurations, the higher the local entropy, implying that the Bayes-optimal solution lays inside a wide flat region. We also consider the algorithmically relevant case of targeting wide flat minima of the (differentiable) mean squared error loss. Our analytical and numerical results show not only that in the balanced case the dependence on the norm of the weights is mild, but also, in the unbalanced case, that the performances can be improved.
\end{abstract}
\maketitle
\tableofcontents{}

\section{Introduction}

Some basic aspects of contemporary machine learning (ML) do not find a satisfactory explanation in  the classical statistical learning framework. For instance, the over-parametrized regime in which deep neural networks (DNN) are utilized does not easily fit into  the uniform convergence scenario in  which one expects that the complexity of a machine learning device (function) should be of the order of the number of examples to provide good generalization properties \cite{shalev2014understanding}.
In order to make progress, researchers are trying to  connect the generalization capabilities of devices such as DNN with the geometrical properties of the error loss functions that is minimized for learning.  
An interesting conjecture which has emerged in various contexts argues that the flatness of the minima can lead to good generalization in the over-parametrized regime~\cite{hochreiter,locentfirst,keskar,jiang2019fantastic,dziugaite1}. The fact that such minima have been recently shown to exist already in shallow networks \cite{locentfirst,baldassi2019shaping,unreasoanable,relu_locent}, puts this conjecture on solid grounds.  Results based on statistical physics techniques, have shown that there exist relatively wide regions with a very high density of optimal minimizers. These regions coexist with a much larger number of critical points (narrow local minima and saddles) and are called either high local entropy regions or wide flat minima.


So far, the theoretical results were limited to the so-called teacher-student scenario in the context of classification: a training set with  i.i.d. randomly-generated inputs and labels provided by a (shallow) teacher network is presented to a student network with the same architecture as the teacher. In the over-parametrized regime, in which the training set does not contain sufficient information, several local minima exist, and the ones with high local entropy were shown to have generalization capabilities close to the Bayesian one.

The over-parametrized teacher-student scenario considered in the above-mentioned studies is highly non-convex, and using random i.i.d inputs is not necessarily realistic. Although it was shown in \cite{baldassi2019shaping} that the phenomenology is similar with real datasets, the problem of obtaining theoretical insight into other classification tasks with different distributions remains open. Some preliminary results in this direction are given in~\cite{Borra_2019, Pastore2020, Rotondo2020, goldt2019modelling,gerace2020generalisation}.

Here, we discuss the connection between local entropy, flatness and generalization in a very basic model of high-dimensional statistical machine learning \cite{mai2019high,Lelarge2019,deng2019model,Lesieur2016}: Gaussian mixtures. The generative model is defined as follows. For a given problem size $N$, an $N$-dimensional vector $\mathbf{v}^\star$ is randomly generated from a standard multivariate normal $\mathcal{N}\left(\mathbf{0},\mathrm{I}_N\right)$. Then we generate a label $\sigma = 1$ or $\sigma=-1$ with probability $\rho$ and $1-\rho$, respectively, and we generate a pattern $\boldsymbol{\xi}$ according to the value of the label $\sigma$ as $\mathcal{N}\left(\sigma \mathbf{v}^\star/\sqrt{N},\Delta \, \mathrm{I}_N\right)$.
In this way we construct a training set with $P \equiv \alpha N$ such patterns; the coordinate $i \in \left\{1,...,N\right\}$ of pattern $\mu$ is therefore given by
\begin{equation}
	\xi_i^\mu = \frac{v_i^\star}{\sqrt{N}} \sigma^\mu + \sqrt{\Delta} \, z_i^\mu
\end{equation}
where $z_i^\mu$ are i.i.d Gaussian random variables with zero mean and unit variance. This results in two potentially overlapping clusters, with the label indicating the cluster a pattern belongs to and where $\Delta$ controls their width. We will refer to the problem with $\rho = 0.5$ as the \emph{balanced} case; we call all other cases \emph{unbalanced}.

As usual in statistical physics, we will consider the high-dimensional limit, where both $N \to \infty$ and $P 
\to \infty$  with the ratio $\alpha = \sfrac{P}{N}$ fixed. 

In this paper we analyze the performance of a threshold-linear classifier (a single-unit neural network). This machine is parametrized by a vector of weights $\boldsymbol{w}$ of length $N$ and a bias $b$, but when studying the balanced case we always simply set $b=0$. The machine predicts the label of a pattern $\boldsymbol{\xi}$ as:
\begin{equation}
    \hat{\sigma}\left(\boldsymbol{\xi};\boldsymbol{w},b\right) =
        \mathrm{sign}\left(
            \frac{1}{\sqrt{N}} \sum_{i=1}^{N} w_i \, \xi_i + b
        \right)
\end{equation}
We can observe that this classifier is invariant to an overall rescaling of the parameters $\boldsymbol{w}^\prime = \kappa \boldsymbol{w}$ and $b^\prime = \kappa b$ for any $\kappa > 0$. It is natural to identify the irrelevant degree of freedom in the parametrization with the norm of $\boldsymbol{w}$.

The most elementary metric by which to measure the performance of this classifier on the training set is the number of errors, which can be expressed as
\begin{equation}
    \mathcal{L}_\mathrm{err}(\boldsymbol{w}, b) =
        \sum_{\mu = 1}^{P} \Theta\left[
            -\sigma^\mu \;
            \hat{\sigma}\!\left(\boldsymbol{\xi}^\mu;\boldsymbol{w},b\right)
        \right]
\end{equation}
where $\Theta\left(\cdot\right)$ is the Heaviside step function, that is $\Theta\left(x\right) = 1$ if $x \geq 0$ and $0$ otherwise. This error-counting loss obviously inherits the scale invariance, but it has the drawback that it cannot be used with the gradient-based methods usually employed in training large neural networks (which is the situation about which we hope to gain the most insight from this simple model). It is therefore of interest to consider a generalized overall loss function form:
\begin{equation}
	\mathcal{L}(\boldsymbol{w}, b) = \sum_{\mu = 1}^{P} \ell \left[ \sigma^\mu \left( \frac{1}{\sqrt{N}} \sum_{i=1}^{N} w_i \, \xi_i^\mu + b \right)  \right] 
\end{equation}
where $\ell\left(\cdot\right)$ is a generic single-pattern loss function. The error-counting case corresponds to $\ell\left(x\right) = \Theta\left(-x\right)$. In what follows we analyze the  mean squared error (MSE) loss $\ell\left(x\right) = \frac{1}{2}\left(x-1\right)^2$, which  is
a well-studied differentiable loss. As for other choices of differentiable losses (e.g. cross-entropy, hinge, etc.) the scale-invariance property is lost, and the role of norm regularizations may become important.

It's important to observe that this model possesses some rather peculiar features if compared to typical classification tasks performed with neural networks, namely the  training loss is convex. Indeed, Bayes-optimal performance can be achieved with a single configuration of the model parameters (instead of requiring a distribution) which can be derived analytically. Additionally, due to the overlap between the Gaussians which are used to generate the data, no classifier can achieve zero test error (in the teacher-student context this would be somewhat similar to the case of having a ``noisy'', unreliable teacher). Therefore, care must be used when considering how the results may generalize to non-convex scenarios.

Recently, this model has been studied in~\cite{Mignacco2020} by using Gordon's inequality. The authors showed that the MSE loss is severely prone to overfitting, especially when $\alpha \simeq 1$\footnote{This value corresponds to the transition point above which the MSE loss has a unique minimum, since minimizing the MSE entails solving a system of $P$ equations in $N$ unknowns; in the balanced case, it is also the transition point where the data is no longer linearly separable.}. They also showed however that, in spite of the fact that the output of the model is norm-independent, the generalization performance is considerably improved by adding to the loss an $L_2$ regularization term on the weights, $\lambda \left\lVert\boldsymbol{w}\right\rVert^2$. For large $N$, the parameter $\lambda>0$ is a Lagrange multiplier that implicitly fixes the norm of the weights. The optimal choice of $\lambda$ depends in general on $\alpha$ and $\rho$. For the balanced case, $\rho=0.5$, the optimum is obtained for all $\alpha$ in the limit $\lambda \to \infty$ (corresponding to vanishing values for the norm of the weights), and in that case the network reaches the Bayes-optimal generalization bound.

In light of these findings, it is interesting to further discuss the role of the norm for these class of  models. As we remarked above, the output of the network (and thus the generalization error) has a scale invariance, i.e. it is independent of the norm (as long as the bias is also properly rescaled). As a consequence we need to understand which are the geometrical features of the solutions space of the classifier that are induced by the regularization of the surrogate loss function used for gradient learning.

It is worth noticing that this scenario also applies to most deep neural network models that use ReLU activations in the intermediate layers and an $\mathrm{argmax}$ operation to produce the output label, and are therefore invariant to uniform scaling of all their weights and biases.
Since the norm cannot affect the generalization capabilities of the network, it seems unlikely that a norm-based regularization could be a valid general strategy.\footnote{There is a caveat to this statement: for particular choices of the loss, e.g.~cross-entropy, it is possible to reparametrize the problem in an invariant way and interpret the norm in terms of a time-evolving parameter of the loss with a ``focusing'' role, see~\cite{baldassi2019shaping}.}

In this paper, we argue for a different, more general criterion to avoid overfitting and improve generalization, proposed in several recent works~\cite{baldassi_local_2016,baldassi2019shaping,relu_locent}, namely that of maximizing the local entropy, which is a particular measure of flatness that can be analyzed theoretically and efficiently approximated algorithmically. We refer to gradient-based algorithms that operate by maximizing (an approximation to) the local entropy as ``entropic algorithms''. One example is given by the replicated stochastic gradient descent (rSGD) algorithm introduced in~\cite{unreasoanable}, where the local entropy is targeted by using several replicas moving in the loss landscape and at the same time feeling an attraction during their dynamics. Another algorithm is entropy-SGD (eSGD)~\cite{entropysgdICLR}, where the local entropy is estimated using stochastic gradient Langevin dynamics~\cite{welling2011bayesian}. Those algorithms have been applied to state of art deep neural networks~\cite{pittorino2020entropic}, proving that they can achieve improved generalization performances.

In particular, for the balanced case, we show analytically that the minimum norm condition, which results in Bayes-optimal performance, corresponds to solutions of maximum local entropy for the classifier (which is norm invariant). We also show that these solutions can be found by entropic algorithms acting on the MSE loss function, and that these algorithms are much less sensitive to the norm.

For the unbalanced case, the authors of~\cite{Mignacco2020} found that, when the bias is learned, reaching the Bayes-optimal generalization error with $L_2$ regularization alone is impossible, and that there exists an optimal finite value of $\lambda$ that minimizes the generalization error. In this paper we show however that that there exists a choice of the bias and of $\lambda$ that allows to reach the Bayes-optimal performance, and that such parameters also have a higher local entropy (measured again in a scale-invariant way). We show both analytically and numerically how to systematically improve the generalization performance in this setting. Learning the bias with entropic algorithms leads to improved performance compared to those which can be attained by the $L_2$ regularized loss function.

The rest of the paper is organized as follows. In section~\ref{sec:TypicalScenario} we briefly review the typical scenario obtained by performing a standard replica-symmetric (RS) replica calculation over the Gibbs measure. We discuss in particular how the choice of the bias is decisive for generalization performances in the unbalanced case. In section~\ref{sec:FranzParisi} we explore the local entropy landscape around the Bayes optimal configuration for the MSE loss function, by performing a calculation \'a la Franz-Parisi~\cite{franz1995recipes}. In section~\ref{sec:ReplicatedCase} we discuss how targeting the local entropy loss we can improve generalization. Finally section~\ref{sec:Conclusions} contains some conclusions.


\section{Bayes-optimal configurations}\label{sec:TypicalScenario}

The partition function of the Gaussian mixture model, with a regularization over the weights of the linear classifier can be written as
\begin{equation}
	\label{eq:partitionfunction}
	Z = \int \prod_i d w_i \, e^{-\beta \mathcal{L}\left(\boldsymbol{w}, b\right) - \frac{\lambda}{2}\sum_i w_i^2}
\end{equation}
where $\beta$ is the inverse temperature. Notice that in our treatment the bias has been fixed as an external parameter. To study the case in which the bias is a learned  parameter we would add another integral over $b$ in the definition of $Z$. 
In the following we will denote the average over the training set with angle brackets:
$\left< \cdot \right> \equiv \prod_{\mu=1}^{P} \mathbb{E}_{\boldsymbol{v}^\star, \sigma^\mu} \mathbb{E}_{\boldsymbol{\xi}^\mu | \boldsymbol{v}^\star, \sigma^\mu} \left[\, \cdot \, \right]$.

\begin{figure}[t]
	\centering
	\includegraphics[width=0.49\textwidth]{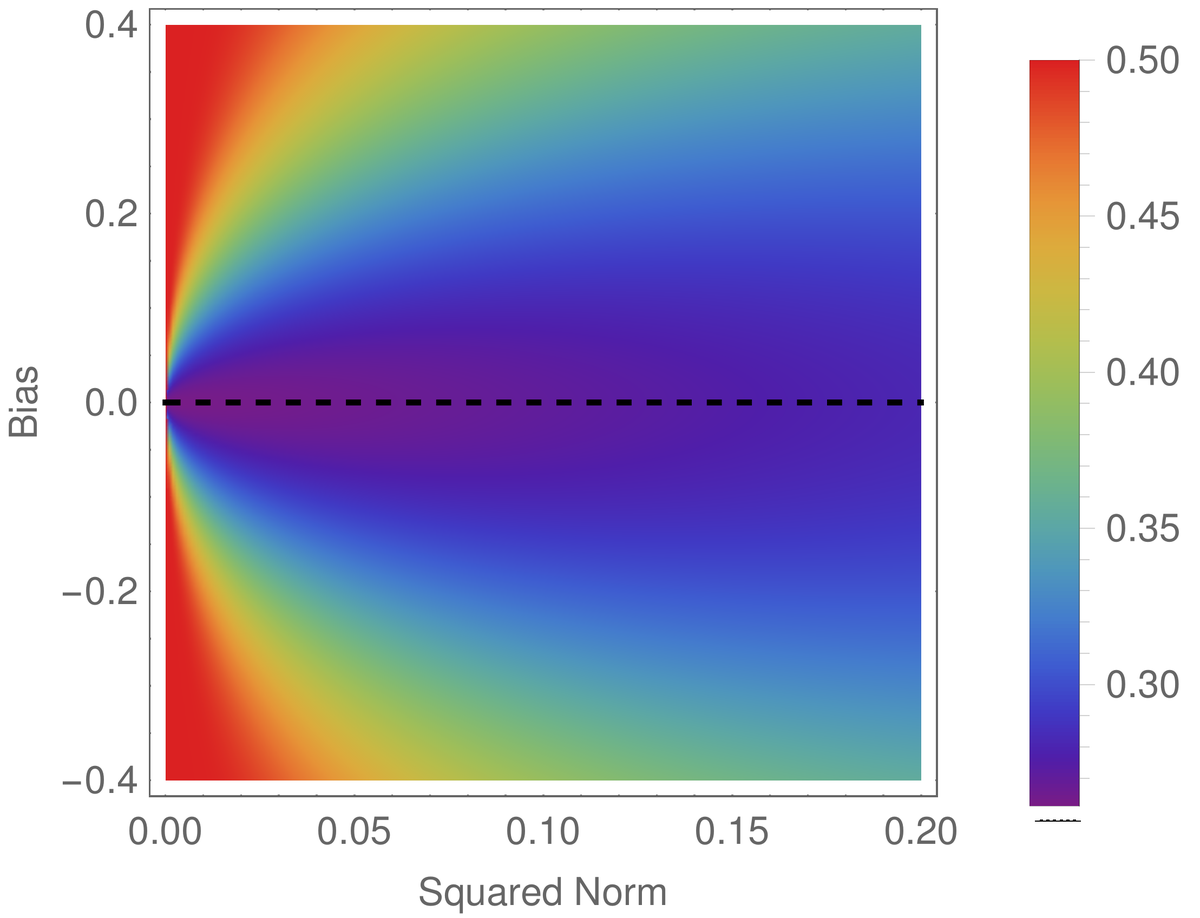}
	\includegraphics[width=0.49\textwidth]{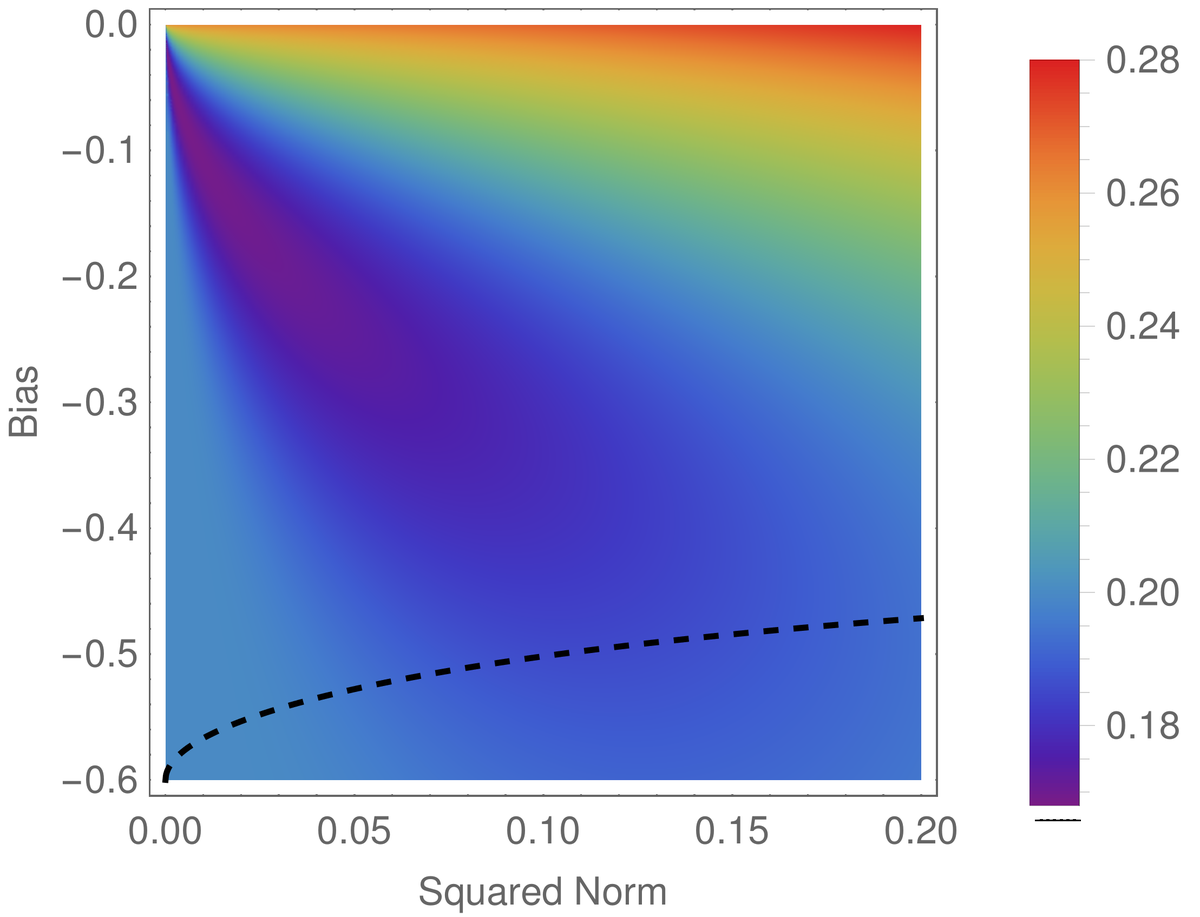}
	\caption{\label{fig:GeneralizationVSbias} Generalization error found by optimizing the regularized MSE loss, as a function of the bias and squared norm. The black dashed line represents the value of the bias obtained by maximizing the Gardner volume. Both plots are for $\alpha = 0.7$ and $\Delta = 1$. Left: Balanced case ($\rho = 0.5$). The Bayes optimal generalization error in this case is $\epsilon_g \simeq 0.2605\dots$ computed using equation~\eqref{eq:BayesGenError} of appendix~\ref{app:BayesOptimal}. Right: Unbalanced case, with $\rho = 0.2$. The Bayes optimal generalization error in this case is $\epsilon_g \simeq 0.1679\dots$. }
\end{figure}

The typical properties of the model are derived by computing the average log-volume $\left<\ln Z\right>/N$, which is the (typical) free entropy of the model $-\beta f$, where $f$ is the corresponding free energy. The free entropy can be computed in the large-$N$ limit using the ``replica trick'':
\begin{equation}
\label{eq:replicatrick}
    \ln Z = \lim_{n \to 0} \partial_n Z^n\,.
\end{equation}
The whole computation in the large $N$ and $\beta$ limit using an RS ansatz is reported in appendix~\ref{app:TypicalCase} and~\ref{app:RS}. Here we discuss the results.
In figure~\ref{fig:GeneralizationVSbias} we show the generalization error found by optimizing the regularized MSE loss function, in the balanced case and one unbalanced case, as a function of the bias and the squared norm (which is implicitly but monotonically controlled by $\lambda$). We also show with a black dashed line the value that the bias takes when it is learned, for any given value of the squared norm. In both the balanced and unbalanced cases there exist choices of the bias and the squared norm that achieve the Bayes optimal performance. However, an important difference can be noted: if we learn the bias in the balanced case we always find $b=0$ for every value of the squared norm; sending $\lambda\to \infty$ (and therefore the squared norm to zero) one recovers the Bayes optimal performance. This is not true in the unbalanced case: fixing $\lambda$ and learning the bias never gives the optimal performance.

\section{The local entropy is larger in the vicinity of the Bayes-optimal configuration}\label{sec:FranzParisi}

In order to quantify the local geometrical landscape around a typical configuration $\Tilde{\boldsymbol{w}}$ of the Gibbs measure with loss function $\mathcal{L}_r  = \sum_\mu \ell_r$, regularization parameter $\lambda_r$ and inverse temperature $\beta_r$, we have studied the so-called Franz-Parisi free entropy~\cite{franz1995recipes,huang2014origin}. It is defined as
\begin{equation}
	\label{eq:FP}
	-\beta f_{\text{FP}}(S) \equiv \frac{1}{N} \left<
	    \frac{
	        \int \prod_i d \Tilde{w}_i \,
	        e^{-\beta_r \mathcal{L}_r\left(\Tilde{\boldsymbol{w}}, \Tilde{b}\right) - \frac{\lambda_r}{2} \sum_i \Tilde{w}_i^2}
	        \ln \mathcal{V}\left(\Tilde{\boldsymbol{w}}, S\right)
	    }{
	        \int \prod_i d \Tilde{w}_i \,
	        e^{-\beta_r \mathcal{L}_r\left(\Tilde{\boldsymbol{w}}, \Tilde{b}\right) - \frac{\lambda_r}{2}  \sum_i \tilde{w}_i^2}
	    }
	\right>
\end{equation}
where the quantity
\begin{equation}
    \label{eq:mV}
	\mathcal{V}\left(\Tilde{\boldsymbol{w}}, S\right) \equiv
	\int d \mu_P(\boldsymbol{w}) \,
	e^{-\beta \mathcal{L}\left(\boldsymbol{w}, \Tilde{b}\right)} \delta\left( \sum_i w_i \tilde w_i - N S\right)
\end{equation}
is the volume of configurations $\boldsymbol{w}$ at inverse temperature $\beta$ that have overlap $S$ with the reference configuration $\Tilde{\boldsymbol{w}}$. The measure $d \mu_P(\boldsymbol{w})$ is a flat measure over a hyper-sphere of squared radius $P$, i.e. the weights $\boldsymbol{w}$ have square norm $\frac{1}{N} \sum_i w_i^2 = P$.  
The overlap $P$ is chosen to match the squared norm of the reference $\Tilde{\boldsymbol{w}}$, that is $P = Q$. Note that $Q$ is fixed via a soft constraint by the regularization parameter $\lambda_r$; in addition we have chosen, for simplicity, the bias of the constrained configuration $\boldsymbol{w}$ to match the one of the reference. 

Notice that in eqs.~\eqref{eq:FP} and~\eqref{eq:mV} we use different losses $\mathcal{L}_r$ and $\mathcal{L}$ (and different parameters too): the landscape of which we explore the geometrical features can differ from the landscape from which we get the reference configuration.

The computation of eq.~\eqref{eq:FP} is long and involved; here we just sketch the main steps, referring to the appendix~\ref{app:FP} for the details. 
The average over the disorder in eq.~\eqref{eq:FP} can be done by using two replica tricks, one for the denominator and another one for the logarithm in the numerator:
\begin{subequations}
    \begin{align}
        \frac{1}{Z} &= \lim_{r \to 0} Z^{r-1} \\
        \ln Z &= \lim_{n\to 0} \partial_n Z^n
    \end{align}
\end{subequations}

Once the average is performed, one has to introduce several order parameters in order to decouple the expressions over the size of the training set $\alpha N$ and of the dimension $N$. Using indexes $a$ or $b$ for replicas in $\left\{1, \dots,  r\right\}$ and $c$, $d \in \left\{1, \dots,  n\right\}$ the order parameters are $p^{cd} = \frac{1}{N} \sum_i w^c_i w^d_i$, $t^{ac} = \frac{1}{N} \sum_i \tilde w^a_i w^c_i$, $O^{c} = \frac{1}{N} \sum_i v_i^\star w^c_i$, $P^{c} = \frac{1}{N} \sum_i (w^c_i)^2$ and the corresponding conjugated ones. Note that $P^c$ is just the squared norm $P$ because of the spherical constraint inside the measure $d\mu_P(\boldsymbol{w})$. Among the conjugated order parameters, we also need to introduce an additional parameter, $\hat S^c$, which imposes the hard constraint on the overlap between the reference configuration $\tilde{\boldsymbol{w}}$ and $\boldsymbol{w}$.

Using an RS ansatz over the order parameters (see appendix~\ref{app:FP_RS}) and performing the large $\beta_r$ limit we obtain
\begin{equation}
\label{eq:FP_free_energy}
	-\beta f_{\text{FP}}\left(S\right) = \mathfrak{G}_S + \alpha \mathfrak{G}_E \,,
\end{equation}
where the definition of the entropic and energetic terms are reported in appendix~\ref{app:FP_lowT}. When $\alpha = 0$ the Franz-Parisi free entropy can be evaluated analytically
\begin{equation}
	\label{eq:FPalpha0}
	-\beta f_{\text{FP}}\left(S, \alpha = 0\right) = \frac{1}{2} \left[ 1 + \ln \left( 2\pi \right) + \ln \left(\frac{1}{\lambda_r}- \lambda_r S^2\right) \right] \,,
\end{equation}
and it gives the logarithm of the total volume of configurations at overlap $S$ with the reference. For a given loss $\ell\left(\cdot\right)$ the local entropy of a given configuration $\tilde{\boldsymbol{w}}$ can be computed by evaluating the local energy $\epsilon_\ell = \frac{\partial \left(\beta f_\text{FP}\right)}{\partial \beta}$ and then using
\begin{equation}
\label{eq:local_entropy}
\mathcal{S} = \beta\left(\epsilon_\ell - f_{\text{FP}}\right)\,.
\end{equation}
The normalized local entropy is just the local entropy~\eqref{eq:local_entropy} minus the total log-volume at $\alpha=0$ given in equation~\eqref{eq:FPalpha0}. It is the logarithm of a fraction of a volume, and thus is upper bounded by zero; additionally, defining the distance as
\begin{equation}
    d \equiv \frac{1}{2} \frac{\sum_{i=1}^N \left(\tilde w_i - w_i\right)^2}{\sum_{i=1}^N \tilde w_i^2} = 1-\frac{S}{P}
\end{equation}
the normalized local entropy is always zero for $d=0$. For $\Tilde{\boldsymbol{w}}$ located in sharp minima  we expect that the normalized local entropy will have a sharp drop near $d \simeq 0$, whereas for flat minima it will be close to zero within some range of small distances.
\begin{figure}[t]
	\centering
	\includegraphics[width=\textwidth]{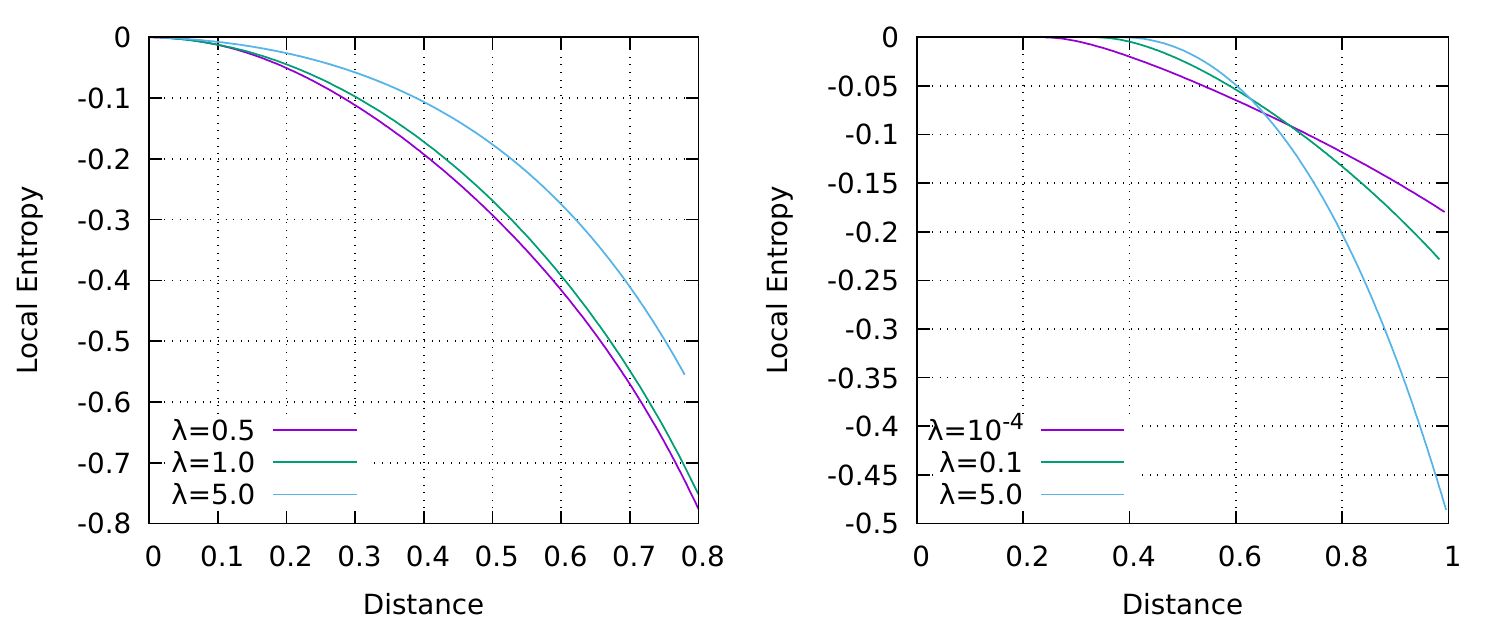}
	\caption{\label{fig:FP_sym} Balanced case ($\rho=0.5$). Normalized local entropy as a function of the distance $d$ computed from reference configurations found by optimizing the regularized MSE loss, with varying regularization strength $\lambda$. Larger values of $\lambda$ correspond to minimizers with better generalization properties. In both figures $\alpha=0.7$, $\Delta=1$, $b=0$. The cutoff $\overline{\epsilon}$ is chosen to be equal to the training error of the reference (left), or is given by the training error of the "oracle" $\boldsymbol{w} = v^\star$ (right panel) as in equation \eqref{eq:train_teacher}.}
\end{figure}

\begin{figure}[t]
	\centering
	\includegraphics[width=\textwidth]{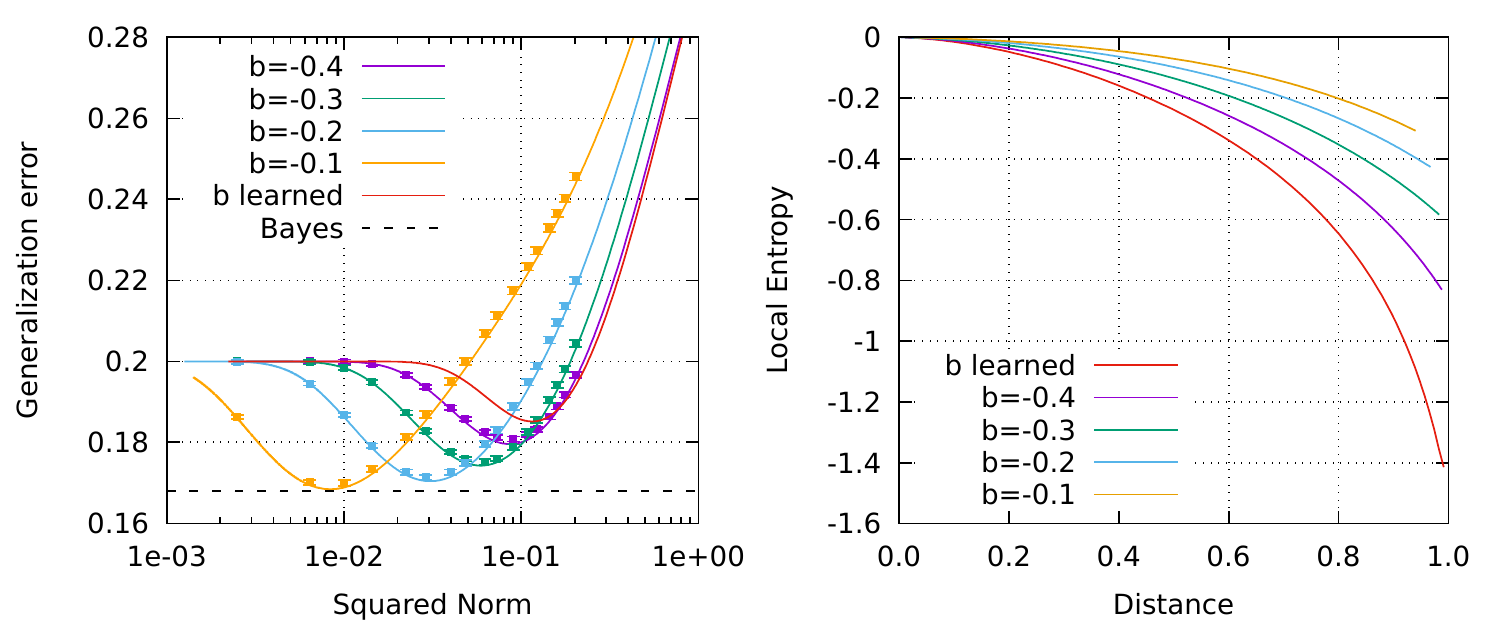}
	\caption{\label{fig:FP_nonsym} Unbalanced case ($\rho=0.2$). In both plots $\alpha = 0.7$ and $\Delta = 1$. Left: Generalization error for a typical minimizer of the MSE loss as a function of the squared norm, for various choices of the bias $b =$ -0.4, -0.3, -0.2, -0.1. Full lines are analytical results, points and error bars are numerical results obtained with $N=1000$. On the red curve, instead, the bias is learned (it's the value that maximizes the free entropy) and thus it's different for every value of the squared norm. The dashed curve is the Bayesian generalization error (see eq.~\eqref{eq:BayesGenError} in the appendix~\ref{app:BayesOptimal}).
	Right: Normalized local entropy as a function of the squared-distance $d$ computed from reference configurations found by optimizing the regularized MSE loss, for various choices of the bias $b_r =$ -0.4, -0.3, -0.2, -0.1 (and $b=b_r$). 
	The corresponding value of the squared norm has been chosen by using the one that minimizes the generalization error for that fixed value of $b$ (see left panel). In the red curve, instead, the bias has been fixed by a saddle point equation (i.e. it is learned). The cutoff $\overline{\epsilon}$ is chosen to be equal to the training error of the reference.}
\end{figure}

We have explored the normalized local entropy landscape of the configurations found by optimizing the regularized MSE loss (i.e. $\ell_r\left(x\right)= \frac{1}{2}\left(x-1\right)^2$), where the \emph{training error} is used in the local entropy definition (i.e. $\ell\left(x\right) = \Theta\left(-x\right)$). We stress that by using the error instead of the MSE we explore the properties of the model in the regime in which it operates during classification.

On the other hand, the parameter $\beta$ has been chosen in such a way that the training error of $\boldsymbol{w}$ given in~\eqref{eq:train_err_FP} is equal to a certain cutoff $\overline{\epsilon}$. 

We have analyzed two different choices for the energy $\overline{\epsilon}$:
\begin{itemize}
	\item in the first case $\overline{\epsilon}$ is chosen to be equal to the training error of the reference (see eq.~\eqref{eq:train_err_ref} in the appendix). This case is depicted in the left panel of figure~\ref{fig:FP_sym} for the balanced case and in the right panel of figure~\ref{fig:FP_nonsym} for the unbalanced case. See the corresponding captions for details.
	
	\item in the second case, only used for the balanced case, $\overline{\epsilon}$ is chosen as the training error of an "oracle classifier" with $\boldsymbol{w} = v^\star$, which is given by:
	\begin{equation}
	\label{eq:train_teacher}
		\begin{split}
			\epsilon_t^\star &=
			\alpha \int \prod_i dv_i^\star d \xi_i
			P\left(\xi_i \mid v_i^\star\right)
			P_v\left(v_i^\star\right) \,
			\theta\left( -\frac{1}{\sqrt{N}} \sum_i v_i^\star \xi_i \right) 
			= \alpha H \left( -\frac{1}{\sqrt{\Delta}} \right)
		\end{split}
	\end{equation}
    This corresponds to the smallest possible test error that any linear classifier machine could achieve, which is non-zero because of the overlap between the two clusters. This case is depicted in the right panel of figure~\ref{fig:FP_sym}.
\end{itemize}

In both cases we clearly see that reference configurations with better generalization properties have higher local entropy curves. We remind that, in the balanced case, the configurations with better generalization properties correspond to larger values of the regularization parameter $\lambda$, whereas in the unbalanced case they correspond to particular fine-tuned values of the bias $b$ and $\lambda$ as already evidenced in the right panel of figure~\ref{fig:GeneralizationVSbias}.

\section{Algorithms that target flatter regions of the MSE landscape also generalize better}\label{sec:ReplicatedCase}

\begin{figure}[t]
	\centering
	\includegraphics[width=\textwidth]{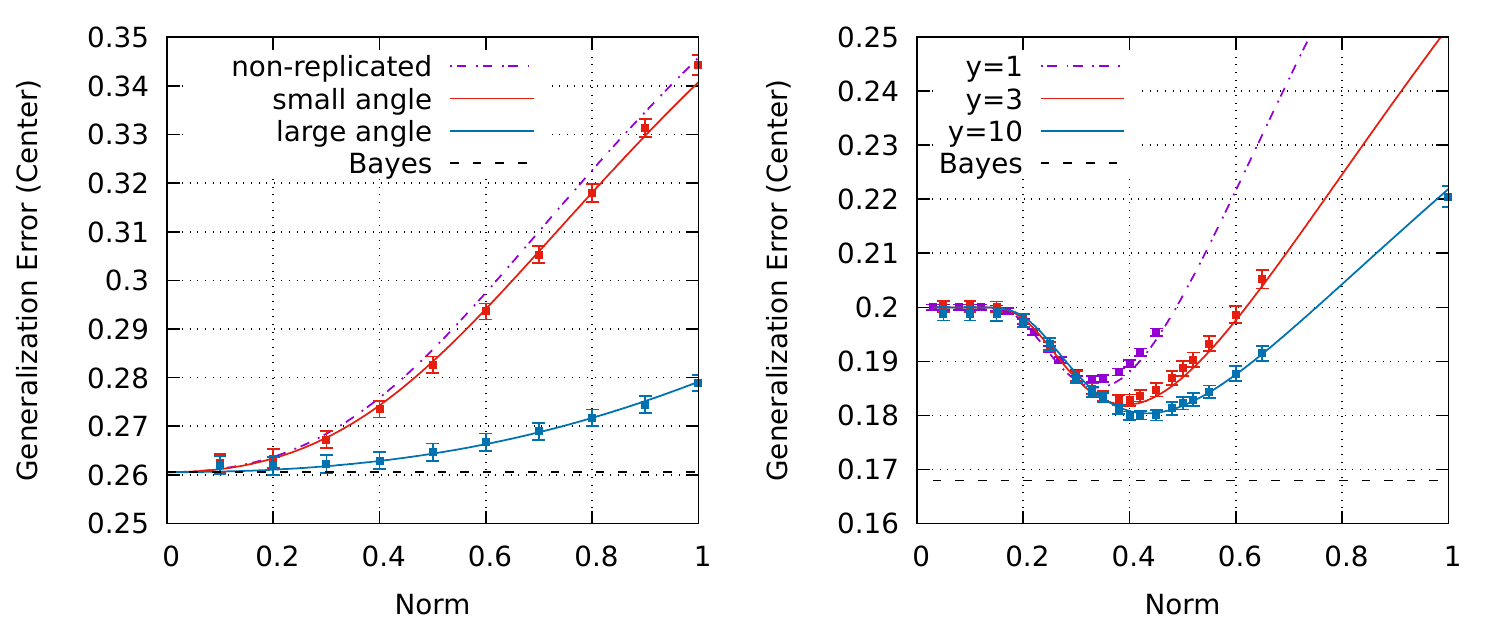}
	\caption{\label{fig:replicated} Generalization error of the center $\bar{\boldsymbol{w}}$ of a system of $y$ replicas, each optimizing the MSE loss and with a constraint on the angle between the replicas, as a function of the norm $n$ of the replicas. In both figures $\alpha=0.7$; the Bayes optimal error is plotted with a dashed black line. Left: Balanced case with $y=10$ replicas. The red curve (small-angle) corresponds to $\cos\left(\theta\right)=0.9$; the large-angle case to $\cos\left(\theta\right)=0.1$. Solid curves are theoretical results, points are numerical results obtained with $N=1000$, averaged over $30$ samples. In the limit $\theta=0$ the results reproduce those of a single device; increasing $\theta$ the dependence on the norm reduces (the curve flattens onto the Bayes-optimal dashed line in the limit $\theta=\pi/2$ and $y\to \infty$). Right: Unbalanced case ($\rho = 0.2$). Solid curves correspond to analytical results, points are numerical results obtained with $N=1000$ and respectively 100, 30, 20 samples for $y=1,3,10$. The angle between the $y$ replicas has been fixed to $\theta = \pi/2$.}
\end{figure}

The results of the previous section confirm that the local entropy landscape constructed using the training error is a good predictor of generalization performance. However, when dealing with much more complex architectures, using the training error as the loss function in eq.~\eqref{eq:local_entropy} is not (yet) algorithmically feasible. In particular, the entropic algorithms rSGD and eSGD must still operate on a differentiable loss. This leaves the question whether targeting high-local-entropy regions in a differentiable loss landscape can still lead to good generalization results open. We have investigated this question analytically on the Gaussian mixture model with a linear classifier and the MSE loss, using the same technique explained in \cite{baldassi2019shaping,relu_locent}, see appendix~\ref{app:1RSB} for details. This amounts to studying the generalization error of the barycenter of a replicated system of $y$ classifiers, each with its own parameters $\boldsymbol{w}^a$ with $a=1,\dots,y$, each optimizing the MSE under constraints on their norms $n$ and on their mutual angles $\theta$, that is: $\forall a,a^\prime:\ \left\lVert \boldsymbol{w}^a\right\rVert  =  n,\ \boldsymbol{w}^a \cdot \boldsymbol{w}^{a^\prime}  =  n^2 \cos\left(\theta\right)$. The barycenter is defined as $\bar{\boldsymbol{w}}= \frac{1}{y}\sum_a \boldsymbol{w}^a$. In this analysis we used the angle $\theta$ rather than the distance in order to compare situations with different norms.\footnote{This is equivalent to using the cosine similarity, often employed in machine learning contexts. We should note however that in a multi-layer classifier the structure of the scale invariance is more complicated and the cosine similarity by itself would not be sufficient to account for it.} 
Notice also that we have not imposed analytically an analogous constraint over the biases of the replicas. In other words, the bias of every replica is the same and we call it $b$. In order to set the bias for the barycenter we recall the fact that the error-counting loss is scale invariant, so that the value of the bias is significant only when compared to the norm of the weights. Additionally we note that in general $\left\lVert \boldsymbol{w}^a\right\rVert\geq\left\lVert \bar{\boldsymbol{w}}\right\rVert$, so if we were to naively use $b$ as bias of the barycenter we would change its relative magnitude respect to $\left\lVert \bar{\boldsymbol{w}}\right\rVert$. For this reason we set $\bar{b}=b \left\Vert\bar{\boldsymbol{w}}\right\Vert / \left\Vert\boldsymbol{w}\right\Vert$.

Our goal is to check if we can improve the generalization performances. Due to the peculiarities of this model, in the balanced case we can simply check whether the barycenter is aligned with the solution of the norm-regularized model with large $\lambda$, which we know to be the optimal classifier.

Some representative results are shown in fig.\ \ref{fig:replicated}.
In the left panel we analyze the balanced case. Our results indicate that, with sufficiently many replicas (even just $y = 3$) and with sufficiently large angles the generalization performance is nearly optimal and the dependence on the norm is mild, and much less pronounced than at small angles (the limit of zero angles reproduces the results of the norm-regularized analysis without replicas). The fact that for this model the best results are obtained with widely separated replicas is due to the simple nature of the problem, and we do not expect this phenomenon to just carry over as-is to the case of deep neural networks, where the landscape is non-convex and the structure of the symmetries is generally much more complex (both in terms of the scale invariance and of discrete permutation symmetries).

In the right panel of figure~\ref{fig:replicated} we show also what happens in the unbalanced case. We have compared the performance of the typical minimizer of the norm-regularized MSE loss with the one of the replicated system. We show that by increasing the number of replicas keeping fixed the angle between them, the generalization performance is improved. The large $y$ limit can be handled analytically, and it is indistinguishable from the results obtained with $y=10$ replicas.

The analytical curves describe very well the numerical results that are obtained with rSGD (see for all the details appendix~\ref{app:numerical_details} and \cite{unreasoanable} where it was firstly introduced). The algorithm consists in training $y$ replicas of a perceptron with and additional term $\mathcal{L}_\mathrm{d}$ in the loss function of each model proportional to the sum of distances from the other replicas; in order to force the replicas to stay at a given distance $d_0$, we modify this term by including $d_0$ as offset:
\begin{equation}
\label{eq:disrance_loss}
\mathcal{L}_\mathrm{d}^{\left(a\right)}=\sum_{b\neq a}^y \left(d_{ab}-d_0\right)^2,
\end{equation}
where the index $a$ refers to the replica of which we are computing the loss. 

\section{Conclusions}\label{sec:Conclusions}
We have presented an analytical study concerning the connection between local entropy and optimal generalization in the case of Gaussian mixtures. Configurations of weights that reach Bayes-optimal performance were shown to be located inside regions of high local entropy, i.e. in wide flat minima of the error-counting loss function. We have also shown that targeting the wide flat minima of the differentiable loss function used for gradient learning (e.g. MSE) is a viable algorithmic strategy. Work is in progress to extend these type of results to deeper architectures.

\section*{Acknowledgments}
\noindent We wish to thank Francesca Mignacco and Luca Saglietti for useful discussions. 

\appendix

\section{The replicated partition function}~\label{app:TypicalCase}
In this first appendix we briefly review how the the geometry of the space of typical Gibbs configurations of the model can be studied using statistical physics techniques~\cite{gardner1988The,gardner1988optimal}, such as the replica method. 

\noindent In the Gaussian mixture problem, the joint probability distribution of patterns, labels and the centroid is 
\begin{equation}
	\label{eq::jpdf_disorder}
	P(\boldsymbol{\xi}, \sigma, \boldsymbol{v}^\star) =  P_\sigma(\sigma) \prod_{i=1}^N \left[ P(\xi_i | \sigma, v_i^\star) P_v(v_i^\star) \right]
\end{equation}
where $P_v(v_i^\star)= \mathcal{N}\left(0, 1 \right)$ and the conditional probability $P(\xi_i | \sigma, v_i^*) = \mathcal{N}\left(\frac{\sigma v_i^\star}{\sqrt{N}}, \Delta \right)$, are Gaussian random variables and $P_\sigma(\sigma) = \rho \delta(\sigma-1) + (1-\rho) \delta(\sigma+1)$.
As stated in the main text, we will denote the average over the disorder distribution given in equation~\eqref{eq::jpdf_disorder} with $\left< \cdot \right> \equiv \prod_{\mu=1}^{P} \mathbb{E}_{\boldsymbol{v}^\star, \sigma^\mu} \mathbb{E}_{\boldsymbol{\xi}^\mu | \boldsymbol{v}^\star, \sigma^\mu} \left[\, \cdot \, \right]$.

The $n-$replicated partition function, which was introduced in~\eqref{eq:replicatrick} to bypass the difficulty of performing the average of the $\log$ of $Z$), can be written as
\begin{multline}
	Z^n = \int \prod_{\mu a} \frac{d h^{\mu a} d \hat h^{\mu a}}{2\pi} e^{i \sum_{\mu a} h^{\mu a} \hat h^{\mu a} -\beta \sum_{\mu a} \ell\left[ \sigma^\mu \left( h^{\mu a} + b \right) \right]} \int \prod_{i a} d W_i^a \, e^{ - \frac{\lambda}{2} \sum_{i a} (W_i^a)^2- \frac{i}{\sqrt{N}}\sum_{\mu a} \hat h^{\mu a} \sum_{i=1}^N W_i^a \xi_i^\mu} \,.
\end{multline}
where as usual we have assumed that $n$ is integer.
In the previous expression we have used a delta function and its integral representation in order to extract the scalar product $\boldsymbol{w}_i \cdot \boldsymbol{\xi}^\mu/\sqrt{N}$ in the loss argument. 
We can now perform the average over the pattern distribution, given the choice of the centroid and label; note that it is factorized over the number of examples in the training set. Indicating by $a$, $b \in \left\{1, \dots, n\right\}$ the replica indexes, we have
\begin{equation}
    \begin{split}
		\prod_{\mu}\left[\mathbb{E}_{\boldsymbol{\xi}^\mu | \boldsymbol{v}^\star, \sigma^\mu} e^{- \frac{i}{\sqrt{N}}\sum_{a} \hat h^{\mu a} \sum_{i} w_i^a \xi_i^\mu} \right] &=  \prod_{\mu} e^{ - \frac{i}{N}\sum_{a} \sigma^\mu \hat h^{\mu a} \sum_{i} W_i^a v_i^* - \frac{\Delta}{2N} \sum_{a} \left(\hat h^{\mu a}\right)^2 \sum_i \left(W_i^a\right)^2 - \frac{\Delta}{N} \sum_{a<b} \hat h^{\mu a} \hat h^{\mu b}  \sum_i W_i^a W_i^b} \\
		&=\int \prod_{a<b} \frac{d q^{ab} d \hat q^{ab}}{2\pi/N} \int \prod_a \frac{d Q^a d \hat Q^a}{2 \pi/N} \frac{d M^a d \hat M^a}{2 \pi/N} e^{-N \sum_{a<b} q^{ab} \hat q^{ab} + \sum_{a<b} \hat q^{ab} \sum_i W_i^a W_i^b}\\
		& \times e^{- N \sum_a Q^a \hat Q^a + \sum_a \hat Q^a \sum_i \left( W_i^a \right)^2 -N \sum_a M^a \hat M^a + \sum_a \hat M^a \sum_i v_i^* W_i^a} \\
		& \times e^{- i \sum_{\mu a} \sigma^\mu \hat h^{\mu a} M^a- \frac{\Delta}{2} \sum_{\mu a} \left( \hat h^{\mu a} \right)^2 Q^a - \Delta \sum_{\mu} \sum_{a<b} q^{ab} \hat h^{\mu a} \hat h^{\mu b} } \,.
		 \end{split}
\end{equation}
In the last equality we have introduced several order parameters
\begin{itemize}
	\item the overlap matrix between two weights $q^{ab} = \frac{1}{N} \sum_i w_i^a w_i^b$ for $a\ne b$;
	\item the squared norm $Q^a = \frac{1}{N}\sum_i (w_i^a)^2$; 
	\item the overlap between a weight and the centroid $M^a = \frac{1}{N} \sum_i v_i^\star w_i^a$;
	\item the corresponding conjugated parameters that we denote by $\hat q^{ab}$, $\hat Q^a$ and $\hat M^a$ respectively that are used to enforce the previous definitions via Dirac delta functions.
\end{itemize}
The final expression for the replicated partition function is
\begin{equation}
	\left<Z^n\right> = \int \prod_{a<b} \frac{d q^{ab} d \hat q^{ab}}{2\pi/N} \int \prod_a \frac{d Q^a d \hat Q^a}{2 \pi/N} \frac{d M^a d \hat M^a}{2 \pi/N}
	e^{-N \sum_{a<b} q^{ab} \hat q^{ab} - N \sum_a Q^a \hat Q^a - N \sum_a M^a \hat M^a + N G_S + N \alpha G_E}
\end{equation}
where we have defined as usual an entropic and an energetic term
\begin{subequations}
	\begin{align}
	G_S &\equiv \ln \mathbb{E}_{v^*} \int \prod_{a} dW^a \, e^{\sum_{a<b} \hat q^{ab} W^a W^b + \sum_a \hat Q^a \left( W^a \right)^2 + v^* \sum_a \hat M^a W^a - \frac{\lambda}{2} \sum_a \left(W^a\right)^2}\\
	G_E &\equiv 
	   \ln \mathbb{E}_{\sigma} \int \prod_{a} \frac{d h^a d \hat h^a}{2\pi} \, e^{- \frac{\Delta}{2} \sum_a \left(\hat h^a\right)^2 Q^a - \Delta \sum_{a<b} q^{ab} \hat h^a \hat h^b + i \sum_a \hat h^a \left( h^a - \sigma M^a  \right) - \beta \sum_a \ell \left[ \sigma \left( h^a + b \right) \right]} 
    \end{align}
\end{subequations}
Note that in the entropic term the average over $v^*$ can be done, so that we obtain
\begin{equation}
	G_S = \ln \int \prod_{a} dW^a \, e^{\sum_{a<b} \left( \hat q^{ab} + \hat M^a \hat M^b \right) W^a W^b  - \frac{1}{2} \sum_a \left( \lambda - \left(\hat M^a \right)^2 - 2\hat Q^a \right) \left(W^a\right)^2} \,.
\end{equation}
As usual in replica computations, we can now employ the saddle point method in order to evaluate the free entropy of the model. Correspondingly, the order parameters satisfy saddle point equations. 
Notice that learning the bias, that is having an additional integral over $b$ in the definition of the partition function~\eqref{eq:partitionfunction}, corresponds to imposing additional saddle point equations for the replicated biases $b^a$, $a=1, \dots, n$. 

\section{Typical case scenario: RS ansatz}\label{app:RS}

In order to study typical solutions, we employ a replica-symmetric (RS) ansatz. This ansatz consists in seeking solutions to the saddle point equations of the form $q^{ab} = q$ for $a \ne b$, $Q^a=Q$, $M^a = M$ and similarly for conjugated order parameters. Defining $\mathcal{G}_S \equiv \lim\limits_{n \to 0} G_S/n$ and $\mathcal{G}_E \equiv \lim\limits_{n \to 0} G_E/n$, we have
\begin{subequations}
	\begin{align}
		\mathcal{G}_S &= \frac{1}{2} \ln \left( \frac{2\pi}{\hat q - 2 \hat Q + \lambda} \right) + \frac{\hat q + \hat M^2}{2 \left( \hat q -2 \hat Q + \lambda \right)}\\
		\label{eq:RSen_term}
		\mathcal{G}_E &= \mathbb{E}_\sigma \int Dx \ln  \int D h \, e^{ - \beta \ell\left( \sqrt{\Delta (Q-q)} h + \sqrt{\Delta q} x + M + \sigma b \right)}
	\end{align}
\end{subequations}
where $Dx \equiv \frac{e^{-x^2/2}}{\sqrt{2\pi}}$ is the standard Gaussian measure. The quenched free entropy $-\beta f \equiv \lim\limits_{n\to 0} \frac{1}{nN}\ln \left<{Z^n}\right>$ is simply
\begin{equation}
	-\beta f = \frac{q \hat q}{2} - Q \hat Q - M \hat M + \mathcal{G}_S + \alpha \mathcal{G}_E
\end{equation}
All other quantities of interest can be computed from the free entropy. For example the training loss is 
\begin{equation}
\label{eq:training_lossRS}
	\epsilon_\ell = \frac{\partial(\beta f)}{\partial \beta} = \alpha \left\langle \ell\left(  \sqrt{\Delta (Q-q)} h + \sqrt{\Delta q} x + M + \sigma b \right) \right \rangle_{x, h}
\end{equation}
where we have defined
\begin{equation}
    \left\langle \mathcal{A}(x, h) \right\rangle_{x, h} \equiv \mathbb{E}_\sigma \int Dx \, \frac{\int Dh \, e^{- \beta \ell\left(  \sqrt{\Delta (Q-q)} h + \sqrt{\Delta q} x + M + \sigma b \right)} \mathcal{A}(x, h) }{ \int Dh \, e^{- \beta \ell\left(  \sqrt{\Delta (Q-q)} h + \sqrt{\Delta q} x + M + \sigma b \right)}}
\end{equation}
The training error corresponding to a typical configuration of the loss $\ell$ is obtained simply by replacing the error counting function in~\eqref{eq:training_lossRS}
\begin{equation}
	\epsilon_t  = \alpha \mathbb{E}_\sigma  \left\langle \Theta\left(-  \sqrt{\Delta (Q-q)} h - \sqrt{\Delta q} x - M - \sigma b \right) \right \rangle_{x, h}
\end{equation}
where $\Theta(\cdot)$ is the Heaviside step function, that is $\Theta(x) = 1$ if $x \geq 0$ and 0 otherwise. 
Other important quantities such as the test loss and the test error are computed and reported in section~\ref{sec:test_loss_error}.

\subsection{Error counting and MSE loss}
Here we specialize the previous expressions in the case of error counting loss $\ell(x) = \Theta(-x)$ and the MSE $\ell(x) = \frac{1}{2}(x-1)^2$ one. In the error counting loss case the energetic term is
\begin{equation}
	\label{eq::G_E_RS_theta}
	\mathcal{G}_E^{\text{err}}  = \mathbb{E}_\sigma \int Dx \ln  H_\beta \left( -\frac{\sqrt{\Delta q} x + M + \sigma b}{\sqrt{\Delta (Q-q)}} \right)
\end{equation}
where $H_\beta(x) \equiv e^{-\beta} + (1-e^{-\beta}) H(x)$ and $H(x) \equiv \int_x^\infty Dh = \frac{1}{2} \text{erfc}\left( \frac{x}{\sqrt{2}} \right)$. Notice that in the error counting loss case the large $\beta$ limit is trivial. 
In the MSE loss all the integrals in the energetic term can be performed, since they are all Gaussians. We obtain
\begin{equation}
	\begin{split}
		\label{eq::G_E_RS_square}
		\mathcal{G}_E^{\text{mse}} = - \frac{1}{2} \ln \left(1+ \beta \Delta(Q-q) \right) - \frac{\beta}{2} \frac{\Delta q + (M-1)^2 +b^2+ 2b(2\rho-1) (M-1)}{1+ \beta \Delta (Q-q)}
	\end{split}
\end{equation}

\subsection{Large $\beta$ limit}

We now concentrate on the case where the loss function has only one minimum. This happens for example in the previously mentioned case of the MSE loss. For those type of losses the overlap tends to the norm with a scaling given by
\begin{equation}
	\label{eq:RS_Scaling_q}
	q = Q - \frac{\delta q}{\beta} \,.
\end{equation}
The conjugated parameters, as a consequence
\begin{subequations}
	\label{eq:RS_Scalings}
	\begin{align}
		\hat q &= \beta^2 \delta \hat Q - \beta \delta \hat q \\
		\hat Q &= \frac{\beta^2}{2} \delta \hat Q - \beta \delta \hat q \\
		\hat M &= \beta \delta \hat M
	\end{align}
\end{subequations}
Notice also that in this limit the interesting regime of the regularization parameter is $\lambda \to \beta \lambda$. 
In this way the free energy in the large $\beta$ limit is 
\begin{equation}
	\begin{split}
		-f = \mathfrak{G}_S + \alpha \mathfrak{G}_E
	\end{split}
\end{equation}
where we have defined
\begin{subequations}
	\begin{align}
		\mathfrak{G}_S &\equiv 
		- \frac{\delta q \delta \hat Q}{2} + \frac{Q \delta \hat q}{2} - M \delta \hat M + \frac{\delta \hat Q + \delta \hat M^2}{2(\lambda + \delta \hat q)} \\
		\mathfrak{G}_E &\equiv 
		- \, \mathbb{E}_\sigma \int D x \, A_\sigma(x)
	\end{align}
\end{subequations}
and $A_\sigma(x) \equiv \min_{h} \left[ \frac{h^2}{2} + \ell\left( \sqrt{\Delta \delta q} h + \sqrt{\Delta Q} x + M + b \sigma \right) \right]$. Calling by $h^*_\sigma(x)$ the corresponding $\text{argmin}$, the saddle point equations are
\begin{subequations}
	\label{eq:SP_RS}
	\begin{align}
		M &= \frac{\delta \hat M}{\lambda + \delta \hat q} \\
		Q &= \frac{\delta \hat Q + \delta \hat M^2}{(\lambda + \delta \hat q)^2} \\
		\delta q &= \frac{1}{\lambda + \delta \hat q} \\
		\delta \hat M &= \frac{\alpha}{\sqrt{\Delta \delta q}} \, \mathbb{E}_\sigma \int Dx \, h^*_\sigma(x) \\
		\delta \hat Q &= \frac{\alpha}{\delta q} \, \mathbb{E}_\sigma \int Dx \, \left[h^*_\sigma(x)\right]^2 \\
		\delta \hat q &= -\frac{\alpha}{\sqrt{Q \delta q }} \,\mathbb{E}_\sigma \int Dx \, x \, h^*_\sigma(x) = -\frac{\alpha}{\sqrt{Q \delta q }} \, \mathbb{E}_\sigma \int Dx \,  \partial_x h^*_\sigma(x)
	\end{align}
\end{subequations}
So far, we have not discussed how to fix the bias. One option is that considered in~\cite{Mignacco2020}, where the bias $b$ is learned in the same way of the weights $w_i$. This corresponds in adding an additional integral over $b$ in the definition of the partition function~\eqref{eq:partitionfunction}. The whole replica analysis can be carried out in a similar way and it leads, in the RS ansatz, to an additional saddle point equation for the bias. This equation is
\begin{equation}
	\label{eq:SP_RS_bias}
	\mathbb{E}_\sigma \, \sigma \int Dx \, h^*_\sigma(x) = 0 \,.
\end{equation} 
Equations~\eqref{eq:SP_RS} together with~\eqref{eq:SP_RS_bias} are the same reported in~\cite{Mignacco2020}, with some change of variables. A second option is to fix the bias simply as an external parameter. 

In the large $\beta$ limit, the training loss becomes
\begin{equation}
	\label{eq:training_loss}
	\epsilon_\ell = \alpha \, \mathbb{E}_\sigma \int Dx \, \ell\left(  \sqrt{\Delta \delta q} \, h^*_\sigma(x) + \sqrt{\Delta Q} x + M + \sigma b \right) \,;
\end{equation}
The training error corresponding to a minimizer of the loss $\ell$, is found, again, by plugging $\ell(x) = \Theta(-x)$ inside~\eqref{eq:training_loss}. In the case of the MSE loss the energetic term simplifies as 
\begin{equation}
	\begin{split}
		\label{eq:G_E_RS_square_betainf}
		\mathfrak{G}_E^{\text{mse}} = - \frac{1}{2} \frac{\Delta Q + (M-1)^2 +b^2+ 2b(2\rho-1) (M-1)}{1+ \beta \Delta \delta q}
	\end{split}
\end{equation}
and the corresponding saddle point equations can be solved analytically as shown in~\cite{Mignacco2020}. For what follows it is important the expression of the training error, which is
\begin{equation}
	\label{eq:train_err_ref}
	\epsilon_t^{\text{mse}} = \alpha \, \mathbb{E}_\sigma H\left( \frac{\Delta \delta q + M + b \sigma}{\sqrt{\Delta Q}} \right) \,,
\end{equation}
The training loss can be easily found by explicitly performing the $y$ integral in~\eqref{eq:training_loss}.
One can verify that when $\lambda$ is increased not only the corresponding squared norm $Q$ lowers, but also, and more importantly, the training error/loss increases (even below the critical capacity $\alpha_c = 1$ of the model, where a zero training error solution can be found). This means that insisting in searching zero error solutions with the MSE loss is counterproductive and leads to overfitting. This is to be expected since the Gaussian mixture model is a particular case of general noisy teacher problems, in which the training set is no longer generated by a rule that can be inferred~\cite{engel-vandenbroek}.

\section{Measuring the local entropy of configurations using the Franz-Parisi approach} \label{app:FP}

In order to compute the average over the disorder induced by the patterns, we use two replica tricks, one for the denominator of~\eqref{eq:FP}, which is just the partition function $\frac{1}{Z} = \lim\limits_{r \to 0} Z^{r-1}$ and one for the log in the numerator of the same equation $\ln Z = \lim\limits_{n \to 0} \partial_n Z^n$.
From now on we will use indexes $a$ or $b$ for replicas in $\left\{1, \dots,  r\right\}$ and $c$, $d \in \left\{1, \dots,  n\right\}$. Therefore we get
\begin{equation}
\begin{split}
	-\beta f_{\text{FP}}(S) &= \frac{1}{N} \lim\limits_{n\to 0} \lim\limits_{r \to 0} \partial_n  \left\langle \int \prod_{a, i} d \Tilde{w}^a_i \prod_a e^{- \beta_r \mathcal{L}_r(\Tilde{\boldsymbol{w}}^a, \Tilde{b}) - \frac{\lambda_r}{2}  \sum_i (\Tilde{w}_i^a)^2} \mathcal{V}^n(\Tilde{\boldsymbol{w}}^{a=1}, S) \right\rangle \\
	&= \frac{1}{N} \lim\limits_{n\to 0} \lim\limits_{r \to 0} \partial_n \left\langle Z_{\text{FP}}^{n,r} \right\rangle\,.
\end{split}
\end{equation}

The computation proceeds as usual by averaging over the disorder and introducing some other order parameters (in addition to those involving only the reference $\Tilde{\boldsymbol{w}}$), namely $p^{cd} = \frac{1}{N} \sum_i w^c_i w^d_i$, $t^{ac} = \frac{1}{N} \sum_i \tilde w^a_i w^c_i$, $O^{c} = \frac{1}{N} \sum_i v_i^\star w^c_i$, $P^{c} = \frac{1}{N} \sum_i (w^c_i)^2$
and the corresponding conjugated ones. Note that $P^c$ is just the squared norm $P$ because of the spherical constraint inside the measure $d\mu_P(\boldsymbol{w})$\footnote{We could have fixed the norm of the of the constrained configuration $\boldsymbol{w}$ by a Lagrange multiplier $\lambda$, in the same way of the reference. The two approaches (microcanonical versus canonical one) are however the same in the thermodynamic limit.}. Among the conjugated order parameters, we need also to introduce an additional parameter, $\hat S^c$, which is the one that impose the hard constraint on the overlap between the reference configuration $\boldsymbol{w}$ and $\boldsymbol{w}$. We finally obtain that the average of $Z_{\text{FP}}^{n,r}$ is
\begin{equation}
	\begin{split}
		\left\langle Z_{\text{FP}}^{n,r} \right\rangle &= \int \prod_{a<b} \frac{d q^{ab} d \hat q^{ab}}{2\pi/N} \int \prod_{c<d} \frac{d p^{cd} d \hat p^{cd}}{2\pi/N} \int \prod_{a>1, c} \frac{d t^{ac} d \hat t^{ac}}{2\pi/N} \int \prod_a \frac{d Q^a d \hat Q^a}{2 \pi/N} \frac{d M^a d \hat M^a}{2 \pi/N}\\
		&\int \prod_c \frac{d \hat P^c}{2 \pi/N} \frac{d O^c d \hat O^c}{2 \pi/N} \frac{d \hat S^c}{2\pi} e^{-N \sum_{a<b} q^{ab} \hat q^{ab} -N \sum_{c<d} p^{cd} \hat p^{cd} - N \sum_{a>1, c} t^{ac} \hat t^{ac}} \\
		& \times e^{- N \sum_a Q^a \hat Q^a - N P\sum_c \hat P^c  -N \sum_a M^a \hat M^a -N \sum_c O^c \hat O^c - NS \sum_c \hat S^c + N G_S + N \alpha G_E}
	\end{split}
\end{equation}
where
\begin{equation}
	\begin{split}
		G_S &\equiv \ln \int \prod_a d \tilde W^a \prod_c d W^c \, e^{\sum_{a<b} (\hat q^{ab} + \hat M^a \hat M^b) \tilde W^a \tilde W^b + \sum_{c<d} (\hat p^{cd} + \hat O^c \hat O^d) W^c W^d} \\
		& \times e^{ \sum_{a>1, c} (\hat t^{ac} + \hat M^a \hat O^c) \tilde W^a W^c - \frac{1}{2} \sum_a ( \lambda_r - (\hat M^a)^2- 2\hat Q^a) (\tilde W^a)^2 }\\
		&\times e^{ \frac{1}{2} \sum_c ((\hat O^c)^2+ 2\hat P^c) (W^c)^2 + \tilde W^{a=1} \sum_c W^c (\hat S^c + \hat M^{a=1} \hat O^c)}
	\end{split}
\end{equation}
and
\begin{equation}
	\begin{split}
		G_E &\equiv \ln \mathbb{E}_\sigma \int \prod_a \frac{d h^a d \hat h^a}{2\pi} \prod_c \frac{d u^c d \hat u^c}{2\pi} e^{- \Delta \sum_{a<b} q^{ab} \hat h^a \hat h^b - \Delta \sum_{c<d} p^{cd} \hat u^c \hat u^d} \\
		& \times e^{ - \Delta \sum_{a>1, c} t^{ac} \hat h^a \hat u^c - S \Delta \hat h^{a=1} \sum_c \hat u^c - \frac{\Delta}{2} \sum_a Q^a (\hat h^a)^2 - \frac{\Delta}{2} P \sum_c (\hat u^c)^2} \\
		& \times e^{ i \sum_a \hat h^a (h^a - \sigma M^a) + i \sum_c \hat u^c (u^c - \sigma O^c) -\beta_r \sum_{a} \ell_r\left[ \sigma \left( h^{a} + \tilde b \right) \right] -\beta \sum_{c} \ell\left[ \sigma \left( u^{c} + b \right) \right] } \,.
	\end{split}
\end{equation}

\subsection{RS ansatz} \label{app:FP_RS}
Next we impose, as usual, an RS ansatz, not only on the order parameters involving the reference as in appendix~\ref{app:RS}, but also on those involving only $\boldsymbol{w}$ or mixing $\boldsymbol{w}$ and $\Tilde{\boldsymbol{w}}$, i.e. $p^{cd} = p$ for $c \ne d$, $t^{ac} = t$ for $a>1$, $O^a=O$, $M^a = M$ and similarly for the conjugated ones. 
We finally get
\begin{equation}
	-\beta f_{\text{FP}}(S) = \frac{p \hat p}{2} + t \hat t - P \hat P - O \hat O - S \hat S + \mathcal{G}_S + \alpha \mathcal{G}_E
\end{equation}
where we have defined the entropic term as
\begin{equation}
	\begin{split}
		\mathcal{G}_S &=
		\frac{1}{\hat p - 2\hat P} \left[ \frac{(\hat S-\hat t)^2\left( 2(\hat q-\hat Q) + \hat M^2+\lambda_r\right)}{2(\hat q -2 \hat Q+ \lambda_r)^2} + \frac{(\hat S-\hat t) (\hat t + \hat M \hat O)}{\hat q - 2\hat Q + \lambda_r}\right] + \frac{\hat p + \hat O^2}{2(\hat p - 2\hat P)}+\frac{1}{2} \ln \left( \frac{2\pi}{\hat p - 2 \hat P} \right) \,.
	\end{split}
\end{equation}
The energetic term instead is
\begin{equation}
	\begin{split}
		\mathcal{G}_E &= \mathbb{E}_\sigma \int Dx \, \frac{1}{\mathcal{Z}(x)} \int Dh \, e^{-\beta_r \ell_r \left( \sigma \tilde b + M + \sqrt{\Delta q} x +\sqrt{\Delta(Q-q)} h \right)} \\
		& \times \int Dy \ln \int Du \, e^{- \beta \ell\left[\sigma b + O + \sqrt{\Delta \gamma - \frac{\Delta (S-t)^2}{Q-q}} y + \frac{\Delta t}{\sqrt{\Delta q}} x + \frac{\Delta(S-t)}{\sqrt{\Delta(Q-q)}} h + \sqrt{\Delta(P-p)} u \right]} \,.
	\end{split}
\end{equation}
In the previous equation have defined $\gamma = p - \frac{t^2}{q}$ and
\begin{equation}
	\mathcal{Z}(x) \equiv \int Dh \, e^{-\beta_r \ell_r \left( \sigma \tilde b + M + \sqrt{\Delta q} x +\sqrt{\Delta(Q-q)} h \right)} \,.
\end{equation}
Note that the parameters involving only the reference $\Tilde{\boldsymbol{w}}$, i.e. $q$, $\hat q$, $\hat Q$, $M$ and $\hat M$ satisfy the same saddle point equations of the previous subsection~\cite{huang2014origin}. 

\subsection{Low temperature limit for the reference configuration} \label{app:FP_lowT}
We are now interested in sending $\beta_r$ to infinity. In order to do that, we need to add to the scalings of the order parameters involving only the reference~\eqref{eq:RS_Scalings}, together with the ones for the overlaps between reference $\Tilde{\boldsymbol{w}}$ and $\boldsymbol{w}$ and their conjugated ones. These are readily seen to be
\begin{subequations}
	\begin{align}
		t &= S - \frac{\delta t}{\beta_r}\\
		\hat t &= \beta_r \, \delta \hat t \\
		\hat S- \hat t &=  \delta \hat S \,.
	\end{align}
\end{subequations}
Using these scalings, the Franz-Parisi entropy is given by equation~\eqref{eq:FP_free_energy}. We have accordingly redefined the entropic and energetic terms, respectively as
\begin{subequations}
	\begin{align}
		\mathfrak{G}_S &= \frac{p \hat p}{2} - \delta t \hat \delta t - P \hat P - O \hat O - S \delta \hat S + \frac{1}{2} \ln \left( \frac{2\pi}{\hat p - 2 \hat P} \right) \nonumber \\
		&+ \frac{1}{\hat p - 2\hat P} \left[ \frac{\hat p + \hat O^2}{2} + \frac{ \delta \hat S^2\left( \delta \hat Q + \delta \hat M^2\right)}{2(\delta \hat q+ \lambda_r)^2} + \frac{\delta \hat S (\delta \hat t+ \delta \hat M \hat O)}{\delta \hat q + \lambda_r}\right] \,,\\
		\mathfrak{G}_E &= \mathbb{E}_\sigma \int Dx Dy \ln \int Du \, e^{- \beta \ell\left[\sigma b + O + \sqrt{\Delta \Gamma} y + \frac{\Delta S}{\sqrt{\Delta Q}} x + \frac{\Delta \delta t}{\sqrt{\Delta \delta q}} h^*_\sigma(x) + \sqrt{\Delta(P-p)} u \right]} \,.
	\end{align}
\end{subequations}
In the last expression we have defined $\Gamma = p - \frac{S^2}{Q}$.

Once $f_\text{FP}$ is known (by solving the corresponding saddle point equations), we can compute the local energy $\epsilon_\ell$ 
as $\epsilon_\ell = \frac{\partial (\beta f_\text{FP})}{\partial \beta}$ and the local entropy as $\mathcal{S} = \beta(\epsilon_\ell - f_{\text{FP}})$. The same formulas are valid if we look to the local entropy landscape in the space of the \emph{training error}, where $\ell(x) = \Theta(-x)$. For clarity we denote the training error as $\epsilon_t$, as in the previous subsection. The training error can be computed by
\begin{equation}
	\label{eq:train_err_FP}
	\epsilon_t 
	= \alpha e^{-\beta} \mathbb{E}_\sigma \int Dx Dy \, \frac{H\left( \frac{\sigma b + O + \sqrt{\Delta \Gamma} y + \frac{\Delta S}{\sqrt{\Delta Q}} x + \frac{\Delta \delta t}{\sqrt{\Delta \delta q}} h^*_\sigma(x)}{ \sqrt{\Delta(P-p)}} \right)}{H_\beta\left( - \frac{\sigma b + O + \sqrt{\Delta \Gamma} y + \frac{\Delta S}{\sqrt{\Delta Q}} x + \frac{\Delta \delta t}{\sqrt{\Delta \delta q}} h^*_\sigma(x)}{ \sqrt{\Delta(P-p)}} \right)}
\end{equation}
where $H_\beta(x) \equiv e^{-\beta} + (1-e^{-\beta}) H(x)$.


\section{Large deviation case: 1RSB ansatz}\label{app:1RSB}

We now study a system of $y$ real replicas where each replica optimizes a loss $\ell$ under constraints on their squared norm $Q$ and also on their mutual angles, namely: $\forall a,b:\ \frac{1}{N}\sum_i (w_i^a)^2 =  Q, \ \frac{1}{N} \sum_i w^a_i w^{b}_i  =  Q \cos\left(\theta\right)$. Thus the partition function of this system of $y$ real replicas  is
\begin{equation}
	\label{eq:replicatedpartitionfunction}
	Z_y = \int \prod_{a} d \mu_Q(\boldsymbol{w}^a) \, e^{-\beta \sum_a \mathcal{L}(\boldsymbol{w}^a, b)} \delta\left(Q \cos\left(\theta\right) - \frac{1}{N} \sum_i w_i^a w_i^b \right)
\end{equation}
Notice that here we are imposing the constraint over the squared norm via an hard constraint (instead of a soft constraint as was used in appendix~\ref{app:RS}). Note, in addition, that the constraint is imposed only on the weights; the biases are assumed to be all equal in all the replicas.
The free entropy of a single replica is $-\beta f = \frac{1}{N y} \overline{\ln Z_y}$ and can be evaluated by the usual replica trick
\begin{equation}
	-\beta f = \lim_{s \to 0} \frac{1}{N y}  \partial_s \overline{Z^{s}}
\end{equation}
It is now evident that if we choose $s=n/y$ virtual replicas, studying a system of $y$ real replicas constrained to be at a mutual overlap $q_1$  
is equivalent to impose a 1RSB ansatz on the standard equilibrium measure~\eqref{eq:partitionfunction} with the Parisi parameter $m$ and the intra-block overlap parameter $q_1$ fixed as external parameters; they play the same role respectively of the number of replicas $y$ and the overlap between replicas $Q^2 \cos(\theta)$ (see also~\cite{monasson1995structural, baldassi2019shaping} for a detailed analysis). The final result is
\begin{equation}
	-\beta f = \frac{q_1 \hat q_1}{2} - \frac{y}{2} \left( q_1 \hat q_1 - q_0 \hat q_0 \right) - Q \hat Q - M \hat M  + \mathcal{G}_S + \alpha \mathcal{G}_E
\end{equation}
where
\begin{subequations}
	\begin{align}
		\mathcal{G}_S &= \frac{1}{2} \ln \left( \frac{2\pi}{\hat q_1 - 2 \hat Q} \right) + \frac{1}{2} \, \frac{\hat q_0 + \hat M^2}{\hat q_1 - 2 \hat Q  - y(\hat q_1- \hat q_0)} + \frac{1}{2y} \ln \left( \frac{\hat q_1 - 2 \hat Q}{\hat q_1 - 2 \hat Q - y(\hat q_1-\hat q_0)} \right)\\
		\mathcal{G}_E &= \frac{1}{y}\mathbb{E}_\sigma \int Dx \, \ln \int Dz \left[ \int D h \, e^{ - \beta \ell\left( \sqrt{\Delta (Q-q_1)} h + \sqrt{\Delta q_0} x + \sqrt{\Delta(q_1-q_0)} z + M + \sigma b \right)} \right]^y  \,.
	\end{align}
\end{subequations}
Notice that in the $q_1 \to Q$ limit, the expressions reduce to the RS case given in eq.~\eqref{eq:RSen_term}, but with $\beta$ replaced by $\beta y$.
Again, in the particular case of the MSE loss, all the integrals in the energetic term can be solved, giving
\begin{equation}
	\label{G_E_square}
	\begin{split}
		\mathcal{G}_E^{\text{mse}} &= - \frac{1}{2} \ln\left( 1+ \beta\Delta(Q-q_1) \right)  + \frac{1}{2y} \ln \left( \frac{1 + \beta \Delta (Q-q_1)}{1+ \beta \Delta (Q-q_1) + y \beta \Delta (q_1-q_0)} \right) \\
		&- \frac{\beta}{2} \frac{\Delta q_0 + \mathbb{E}_\sigma (M+b\sigma - 1)^2}{1+ \beta \Delta (Q-q_1) + \beta y \Delta(q_1-q_0)} \,.
	\end{split}
\end{equation}

\subsection{Computing the barycenter of the replicas}

From the system of $y$ real replicas one can compute new valid configurations of the weights that can achieve good generalization properties. As shown in~\cite{locentfirst} one of those is the barycenter of the replicated system, defined as
\begin{equation}
	\overline{w}_i \equiv \frac{1}{y}\sum_{a} w_i^a \,.
\end{equation}
The barycenter is identified by its overlap with the teacher $\overline{M} = \frac{1}{N} \sum_i \overline{w}_i v_i^\star$ and its squared norm $\overline{Q} = \frac{1}{N} \sum_i \overline{w}_i^2$. They can computed by expressing them in terms of the known replica-overlap quantities. In fact
\begin{subequations}
	\begin{align}
		\overline{M} &= \frac{1}{N y} \sum_i \sum_{a} w_i^a v_i^\star = \frac{1}{y}\sum_a M^a\\
		\overline{Q} &= \frac{1}{N} \sum_i  \overline{w}_i^2 = \frac{1}{y^2 N} \sum_{a b} \sum_i  w_i^a w_i^b = \frac{1}{y^2} \sum_a Q^a + \frac{1}{y^2} \sum_{a \ne b} q^{ab}	
	\end{align}
\end{subequations}
Since all real replicas are have the same mutual overlap and the same squared norm, we get
\begin{subequations}
	\begin{align}
		\overline{M} &= M \\
		\overline{Q} &= \frac{Q-q_1}{y} +q_1
	\end{align}
\end{subequations}
Notice that since $|q_1|<Q$, the squared norm of the center is always lower than that of a real replica. In addition, there is always a maximal angle at which the replicas can be placed. This value is
\begin{equation}
	\label{eq:thetamax}
	\theta_\text{max} = \arccos\left( - \frac{1}{y-1} \right)
\end{equation}
for example, when $y=3$, the maximal angle is clearly $\theta_\text{max} = \frac{2\pi}{3}$, i.e. the replicas are placed on the vertices of a equilateral triangle. In general at the maximal angle the $y$ replicas are placed on the vertices of a regular $y$-simplex\footnote{Since in an $y$-simplex $\sum_{a=1}^y \boldsymbol{u}^{a}=0$ where $||\boldsymbol{u}^a||=1$ are unit-norm vectors that identify the position of the vertices with respect to the barycenter, we have $|| \sum_{a=1}^y \boldsymbol{u}^{a}||^2 =  y + y(y-1) \cos(\theta) = 0$ which gives eq.~\eqref{eq:thetamax}.} immersed on the $(N-1)$-sphere of radius $\sqrt{N}$.

\subsection{Large-$\beta$ limit for the MSE loss}

Let us now perform the large $\beta$ limit in the case of a loss having only one minimum. We restrict for simplicity to the MSE loss case. 

It can be seen that the correct scaling of $q_0$ is
\begin{equation}
	q_0 = \frac{Q-q_1}{y} + q_1 - \frac{\delta q_0}{\beta} \,.
\end{equation}
and the conjugated parameters instead scale as
\begin{subequations}
	\begin{align}
		\hat q_0 &= \beta^2 \delta \hat q_0 + \frac{\beta}{2} \delta \hat q_1 \\
		\hat q_1 &= \beta^2 \delta \hat q_0 - \frac{\beta}{2} \delta \hat q_1 \\
		\hat Q &=  \frac{\hat q_1}{2} - \frac{1}{2(Q-q_1)} \\
		\hat M &= \beta \delta \hat M 
	\end{align}
\end{subequations}
The free energy is therefore
\begin{equation}
	- f = \frac{1}{2}(Q-q_1)\delta \hat q_1 + \frac{y}{2} \left[ q_1 \delta \hat q_1 - \delta q_0  \delta \hat q_0 \right] - M \delta \hat M   + \mathfrak{G}_S + \alpha \mathfrak{G}_E
\end{equation}
where the new entropic and energetic terms (rescaled with $\beta$) are
\begin{subequations}
	\begin{align}
		\mathfrak{G}_S^{\text{mse}} &\equiv \lim\limits_{\beta \to \infty}\frac{\mathcal{G}_S^{\text{mse}}}{\beta} = \frac{1}{2} \, \frac{ \delta \hat q_0 + \delta \hat M^2}{ y \delta \hat q_1}\\
		\mathfrak{G}_E^{\text{mse}} &\equiv \lim\limits_{\beta \to \infty}\frac{\mathcal{G}_E^{\text{mse}}}{\beta} = - \frac{1}{2} \frac{\Delta \left(\frac{Q-q_1}{y} + q_1\right) + \mathbb{E}_\sigma (M+b\sigma - 1)^2}{1 + y\Delta \delta q_0}
	\end{align}
\end{subequations}
The training error is
\begin{equation}
	\epsilon_t^{\text{mse}} = \alpha \mathbb{E}_\sigma H\left( \frac{\Delta \delta q_0 + M + b \sigma}{\Delta \left( \frac{Q-q_1}{y}+q_1 \right)} \right)
\end{equation}
Notice that in the large $\beta$ limit the training error/loss of one of the replicas is not zero, because of distance constraint and the convexity of the loss landscape.

The large $y$-limit can also be handled (in a way similar to the case of a generic loss): it suffices to rescale the order parameters $\delta q_0 \to \sfrac{\delta q_0}{y}$ and $\delta \hat q_1 \to \sfrac{\delta \hat q_1}{y}$.

\section{Test loss and generalization error}
\label{sec:test_loss_error}
Let us compute the test loss of a configuration of weights $\boldsymbol{w}$ that has overlap with the teacher $M$ and squared norm $Q$ after a training with $\alpha N$ patterns. In order to do that, we need to present to $\boldsymbol{w}$ a new example that we will denote by $\boldsymbol{\xi}$ with its corresponding label $\sigma$. The test loss is therefore
\begin{equation}
\begin{split}
	\mathcal{L}_t(\boldsymbol{w}, \boldsymbol{v}^\star) &\equiv \mathbb{E}_\sigma \mathbb{E}_{\boldsymbol{\xi} | \boldsymbol{v}^\star, \sigma} \, \left\langle \ell\left[ \sigma \left( \frac{1}{\sqrt{N}} \sum_{i=1}^{N} w_i \, \xi_i + b \right) \right] \right\rangle_{Z} 
\end{split}
\end{equation}
where $\langle \bullet \rangle_Z$ is the average over $\boldsymbol{w}$ extracted according to the partition function given in~\eqref{eq:partitionfunction}. 
Extracting the loss argument by a delta function and performing the expectation over the pattern distribution, we finally get
\begin{equation}
	\label{test_loss}
	\mathcal{L}_t(\boldsymbol{w}, \boldsymbol{v}^\star)  = 
	\mathbb{E}_\sigma \int Du \, \ell\left(  \sigma b + M + \sqrt{\Delta Q} u \right)\,.
\end{equation}
In the case of the error counting loss $\ell(x) = \Theta(-x)$ we get the expression for the generalization error given in~\cite{Mignacco2020} 
\begin{equation}
	\label{gen_theta}
	\epsilon_g = \rho H\left( \frac{M + b}{\sqrt{\Delta Q}} \right) + (1-\rho) H\left( \frac{M - b}{\sqrt{\Delta Q}} \right) \,.
\end{equation}
For the MSE loss instead we obtain
\begin{equation}
	\label{test_loss_theta}
	\mathcal{L}^{\text{mse}}_t(\boldsymbol{w}, \boldsymbol{v}^\star) = 
	\frac{1}{2} \left[ \Delta Q + \mathbb{E}_\sigma  (M-1 + \sigma b)^2 \right]\,.
\end{equation}

\section{Bayesian generalization error}\label{app:BayesOptimal}

In this section we derive the expression of the Bayesian generalization error. For a given generator $\boldsymbol{v}^\star$ and a configuration of the weights $\boldsymbol{w}$ the generalization error is found by extracting a test pattern $\boldsymbol{\xi}$ with label $\sigma$ and computing the corresponding error
\begin{equation}
\begin{split}
	\epsilon_g(\boldsymbol{w}, \boldsymbol{v}^\star) &= \mathbb{E}_\sigma\int d \boldsymbol{\xi} \, P(\boldsymbol{\xi}, \sigma | \boldsymbol{v}^\star) \Theta\left(-\sigma \hat \sigma(\boldsymbol{\xi}; \boldsymbol{w}, b) \right)
	\\ 
	&=\mathbb{E}_\sigma \int Du \, \Theta\left( - \sigma b - \frac{1}{N} \sum_i w_i v_i^\star - \sqrt{\frac{\Delta}{N} \sum_i w_i^2} \, u \right) \\
	&= \mathbb{E}_\sigma H\left( \frac{\sigma b + \frac{1}{N} \sum_i w_i v_i^\star}{\sqrt{\frac{\Delta}{N} \sum_i w_i^2} }\right) \,.
\end{split}
\end{equation}
The Bayesian mean generalization error is obtained by averaging $\epsilon_g(\boldsymbol{w}, \boldsymbol{v}^\star)$ with respect to the posterior $P(\boldsymbol{v} | \left\{ \boldsymbol{\xi}^\mu, \sigma^\mu \right\})$. Applying Bayes theorem we have
\begin{equation}
	\label{Bayesian_mean_loss}
	\begin{split}
		\epsilon_g^B(\boldsymbol{w} | \left\{ \boldsymbol{\xi}^\mu , \sigma^\mu \right\}) &= \int d \boldsymbol{v}^\star \, P(\boldsymbol{v}^\star | \left\{ \boldsymbol{\xi}^\mu, \sigma^\mu \right\}) \epsilon_{g}(\boldsymbol{w}, \boldsymbol{v}^\star) = \frac{\int d  \boldsymbol{v}^\star \, P(\left\{ \boldsymbol{\xi}^\mu, \sigma^\mu \right\} |  \boldsymbol{v}^\star) P( \boldsymbol{v}^\star) \, \epsilon_{g}(\boldsymbol{w}, \boldsymbol{v}^\star)}{\int d  \boldsymbol{v}^\star \, P(\left\{ \boldsymbol{\xi}^\mu, \sigma^\mu \right\} |  \boldsymbol{v}^\star) P( \boldsymbol{v}^\star) }  \\
		& = \mathbb{E}_\sigma H\left(\frac{\sigma b + \frac{1}{N} \sum_i w_i \overline{v}_i}{\sqrt{\frac{\Delta}{N} \sum_i w_i^2}}\right) \,,
	\end{split}
\end{equation}
where we have defined
\begin{equation}
	\overline{v}_i \equiv \frac{1}{\Delta + \alpha} \frac{\sum_\mu \xi_i^\mu \sigma^\mu}{\sqrt{N}} \,.
\end{equation}
Minimizing~\eqref{Bayesian_mean_loss} with respect to $\boldsymbol{w}$ and $b$, we obtain
\begin{subequations}
    \begin{align}
        \boldsymbol{w}^{\text{opt}}(\left\{\boldsymbol{\xi}^\mu, \sigma^\mu\right\}) &= \overline{\boldsymbol{v}} \\
        b^{\text{opt}} &= \frac{\Delta}{2} \ln\left(\frac{\rho}{1-\rho} \right) \,.
    \end{align}
\end{subequations}
We can now use $\boldsymbol{w}^{\text{opt}}$ and $b^{\text{opt}}$ to compute the minimal expected generalization error: we first extract a $\boldsymbol{v}^\star$, with that a training set $\left\{ \boldsymbol{\xi}^\mu, \sigma^\mu \right\}$, from that compute $\boldsymbol{w}^{\text{opt}}$ and $b^{\text{opt}}$, then extract a test pattern $\boldsymbol{\xi}^\star$ with label $\sigma^\star$ and finally use it to compute the error. We have
\begin{equation}
		\epsilon_g^B = \int d \boldsymbol{v}^\star \, d \boldsymbol{\xi}^\star \, d \sigma^\star \,  P(\boldsymbol{\xi}^\star, \sigma^\star | \boldsymbol{v}^\star) \int \prod_{\mu} d\boldsymbol{\xi}^\mu d \sigma^\mu \prod_{\mu} P(\boldsymbol{\xi}^\mu, \sigma^\mu | \boldsymbol{v}^\star) P(\boldsymbol{v}^\star) \, \Theta\left(-\sigma^\star \hat \sigma(\boldsymbol{\xi}^\star; \boldsymbol{w}^{\text{opt}}, b^{\text{opt}}) \right)  \,.
\end{equation}
By using the central limit theorem repeatedly, we obtain
\begin{equation}
\label{eq:BayesGenError}
		\epsilon_g^B =  \mathbb{E}_{\sigma^\star} H\left( \frac{M^{\text{opt}} + b^{\text{opt}} \sigma^\star}{\sqrt{\Delta Q^{\text{opt}}}} \right) \,,
\end{equation}
where we have defined 
\begin{equation}
		M^{\text{opt}} = Q^{\text{opt}} \equiv \frac{\alpha}{\Delta+\alpha} \,.
\end{equation}

\clearpage
\section{Numerical details}~\label{app:numerical_details}

We used rSGD for all simulations reported in this work. As described in the main text, the algorithm consists in training $y$ replicas of a perceptron each initialized in a different way, with an additional term in the loss function of each model proportional to the sum of distances from the other replicas. The total loss function is
\begin{equation}
\label{eq:total_loss}
	\begin{split}
    \mathcal{L}(\{\boldsymbol{w}^a,b^a\}_{a=1}^y)&=\sum_{a=1}^{y}\left[\mathcal{L}^a_{\mathrm{MSE}}+\lambda \mathcal{L}_\mathrm{d}^{a}\right] \\
    &=\sum_{a=1}^{y}\left[\mathcal{L}^a_{\mathrm{MSE}}+\lambda\sum_{a\neq b}^y \left(d_{ab}-d_0\right)^2\right] \\
    \end{split}
\end{equation}
where $\mathcal{L}_\mathrm{d}^{a}$ is the term we introduced in order to force the replicas to stay at a given distance $d_0$. The bias is treated separately: in the cases where $\rho=0.5$ it is simply set to zero; in the cases where $\rho\neq 0.5$ we add to the loss of each replica $\lambda\sum_{a\neq b}^y \left(b_{a}-b_b\right)^2$. This is done in order to match results with analytical calculations, where the replicas share the same bias. 

Then we define a center model as our predictor, defined as the average of the replicas in the following way:
\begin{subequations}
	\label{eq:center_model}
	\begin{align}
		\bar{\boldsymbol{w}} &= \frac{1}{y}\sum_{a=1}^y \boldsymbol{w}^a\\
		\bar{b} &= \left\Vert\bar{\boldsymbol{w}}\right\Vert\frac{1}{y}\sum_{a=1}^y \frac{b^a}{\left\Vert\boldsymbol{w}^a\right\Vert}
	\end{align}
\end{subequations}
The perceptron $\left(\bar{\boldsymbol{w}},\bar{b}\right)$ is the model we use to compute loss and error on both the testset and the trainset. Note that, as discussed in the main text, the bias of the center model is not simply the average of the biases, but rather the average of the biases weighted by the inverse norm of the weights, scaled by the norm of the predictor itself. Note that at the end of the training the replicas are expected to have the same value $b$ of the bias, and since the norm of the replicas is fixed to some given $\left\Vert\boldsymbol{w}\right\Vert$ the bias of the center will simply be $\bar{b}=b \left\Vert\bar{\boldsymbol{w}}\right\Vert / \left\Vert\boldsymbol{w}\right\Vert$.

Most of the following details should not matter for the sake of generalization error because the problem is convex. They still determine the rate of convergence to the analytical solution, so we report them in detail.

The training is performed with PyTorch by using full-batch gradient descent with learning rate $1\cdot10^{-4}$. Initialization is standard Xavier. In the cases with $y=1$ we train with the Adam optimizer for $2\cdot10^4$ epochs. 

In the cases with $y>1$ we train with the SGD optimizer with momentum 0.5 for $4\cdot10^4$ epochs. In those cases we increase the coupling constant $\lambda$ at each epoch by a factor $\lambda_1=5\cdot10^{-3}$ starting from the value $\lambda_0=1\cdot10^{-4}$ up to a maximum $\lambda_{\mathrm{max}}=1\cdot10^2$; namely we set $\lambda(t)=\min[\lambda_0(1+\lambda_1)^t,\lambda_{\mathrm{max}}]$.

The norm is always kept fixed by renormalizing the weights to the given magnitude before each forward pass of the perceptron. 

\bibliography{references.bib}
\end{document}